\documentclass[conference]{IEEEtran}
\IEEEoverridecommandlockouts
\usepackage{cite}
\usepackage{amsmath,amssymb,amsfonts}
\usepackage{algorithmic}
\usepackage{graphicx}
\usepackage{textcomp}
\usepackage{xcolor}
\usepackage{balance}
\usepackage{fancyhdr}
\def\BibTeX{{\rm B\kern-.05em{\sc i\kern-.025em b}\kern-.08em
    T\kern-.1667em\lower.7ex\hbox{E}\kern-.125emX}}

\graphicspath{ {./images/} }

\begin{document}

\title{Facial Landmark Predictions with Applications to Metaverse}

\author{\IEEEauthorblockN{Qiao Han}
\IEEEauthorblockA{\textit{Computer Science and Engineering} \\
\textit{Nanyang Technological University}\\
Singapore, Singapore \\
qiao002@e.ntu.edu.sg}
\and
\IEEEauthorblockN{Jun Zhao}
\IEEEauthorblockA{\textit{Computer Science and Engineering} \\
\textit{Nanyang Technological University}\\
Singapore, Singapore \\
junzhao@ntu.edu.sg}
\and
\IEEEauthorblockN{Kwok-Yan Lam}
\IEEEauthorblockA{\textit{Computer Science and Engineering} \\
\textit{Nanyang Technological University}\\
Singapore, Singapore \\
kwokyan.lam@ntu.edu.sg}
}

\maketitle
\thispagestyle{fancy}
\pagestyle{fancy}
\lhead{This paper appears in the Proceedings of 2022 IEEE 8th World Forum on Internet of Things (WF-IoT).\\ Please feel free to contact us for questions or remarks.}
\cfoot{~\\[-25pt]\thepage}

\begin{abstract}
This research aims to make metaverse characters more realistic by adding lip animations learnt from videos in the wild. To achieve this, our approach\footnote{Code is available at https://github.com/sweatybridge/text-to-anime} is to extend Tacotron 2 text-to-speech synthesizer to generate lip movements together with mel spectrogram in one pass. The encoder and gate layer weights are pre-trained on LJ Speech 1.1 data set while the decoder is retrained on 93 clips of TED talk videos extracted from LRS 3 data set. Our novel decoder predicts displacement in 20 lip landmark positions across time, using labels automatically extracted by OpenFace 2.0 landmark predictor. Training converged in 7 hours using less than 5 minutes of video. We conducted ablation study for Pre/Post-Net and pre-trained encoder weights to demonstrate the effectiveness of transfer learning between audio and visual speech data.
\end{abstract}


\section{Introduction}

The relative importance of words, tone of voice, and body language in communications, known as the 7/38/55 rule, was published by Albert Mehrabian in 1967. While the studies have several limitations, it highlights the effectiveness of non-verbal cues in conversations as 55\% of your message comes from your body language. Applying this finding to the domain of text-to-speech synthesis, we believe adding animated lip movements could make computer generated speech easier to understand. For the metaverse, better lip sync means more natural looking avatars \cite{vrlips}. And for people with speech disabilities, this could help them better communicate in real life through both synthetic speech and visual appearance \cite{disable}.

Prior to deep learning, traditional text-to-speech programs relied heavily on hand engineered acoustic features as input \cite{classic}. This required a lot of domain expertise to curate training data and hence limited the model's ability to learn. Furthermore, the synthetic output of these models was usually fitted to a parametric face model used by specific animators. This further constrained learning to pre-programmed sequences without accounting for all variations in real life.

\begin{figure}[t]
\centering
\includegraphics[width=\linewidth]{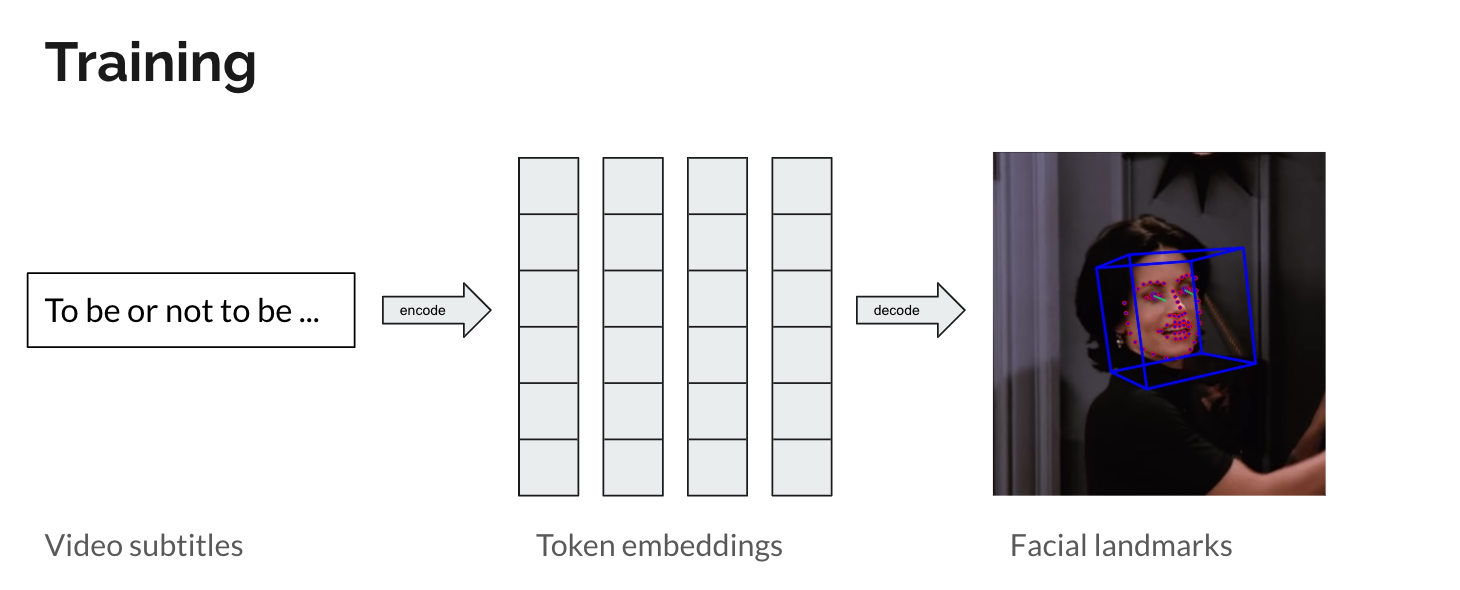}
\includegraphics[width=\linewidth]{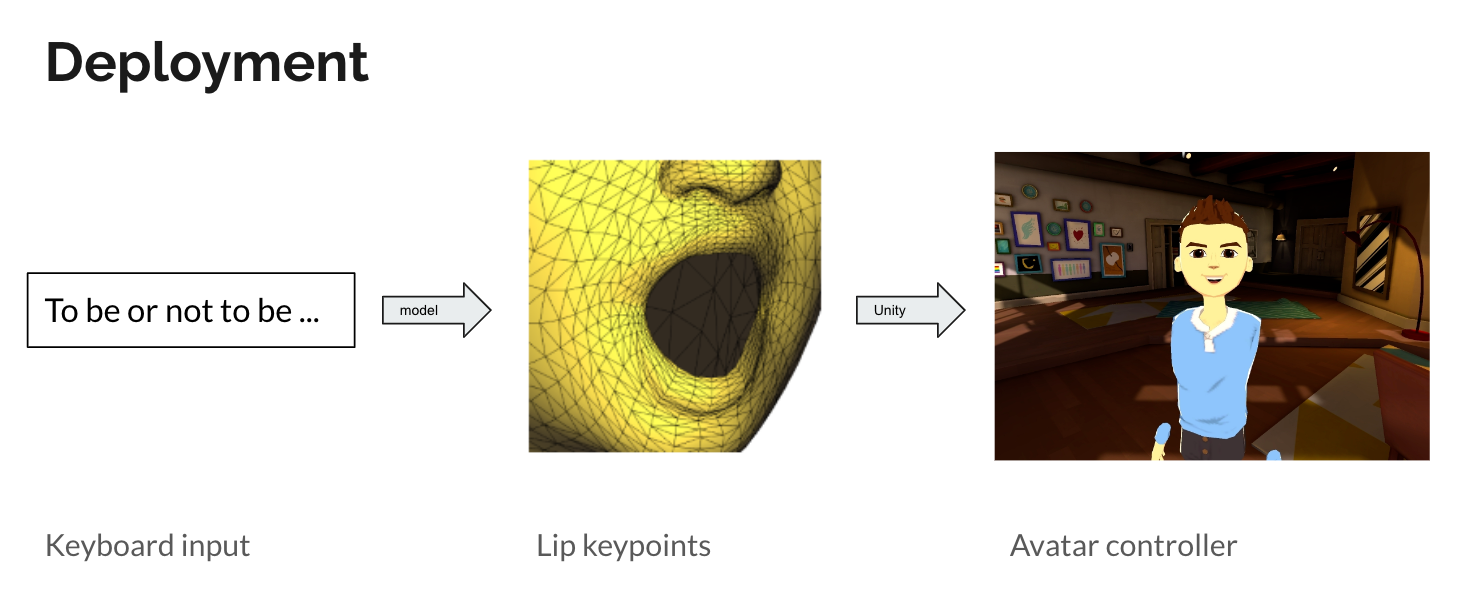}
\caption{\textbf{Training:} model learns to generate lip movements from videos in the wild by mapping time-synced subtitles to video frames. \textbf{Deployment:} metaverse clients receive text input over the network. Model predicts lip keypoints in real-time to animate user avatars.}
\label{fig:concept}
\end{figure}

In this paper, we extend Nvidia's open source implementation of Tacotron 2 \cite{tacotron} neural network to learn directly from text embedding and facial landmarks in raw videos (Fig. \ref{fig:concept}). The model predicts the movement trajectory of 20 facial landmarks representing the lips as the speaker as he or she pronounces each syllable. When deploying for metaverse applications, we envision the model can take both player chat bubbles and NPC dialog as input to synthesize realistic avatar lip movements.

Using just 5 minutes of transcribed video clips as training data, our model learns precise 3D locations of a speaker's lips with 8mm average error from the ground truth. As our data labelling process is fully automated using OpenFace 2.0 \cite{openface} keypoint detection model, we can potentially scale up the training set to thousands of unseen videos.

\section{Related Work}

In metaverse, the pre-dominant approach to lip sync is audio-driven \cite{language,real}. For example, Karras ~\textit{et al.}~ \cite{audio2face} used 3-5 minutes of animation data to train a neural network capable of predicting a fixed-topology head mesh from half a second of mel-spectrogram. Richard ~\textit{et al.}~ \cite{gaze} collected 5 hours of high frame rate 3D face scans as training data for animating codec avatars. Eskimez, Maddox, Xu, and Duan \cite{generate,gen3d} took the first and second-order temporal differences of mel-spectrogram to predict 3D facial landmarks. We differ from their approaches in that we use text embedding as model input. Since our approach is text-driven, it avoids delays in audio signal arrival and is closer to how speech is generated by humans.

Another related line of work is to generate photo realistic talking heads from text input \cite{express,old_cam,old_ms,coupled}. The key idea is to sample a learned Hidden Markov Model at each time step. The sampled parameters are passed to Active Appearance Models for image generation. For industrial applications, Zhou ~\textit{et al.}~ \cite{viseme} explored generating compact viseme curves for JALI or other FACS based face models. Instead of using parametric models, our approach predicts 3D vertex positions directly.

It is also worth noting that the formulation of our problem is the inverse of automatic speech recognition or lip reading. Instead of decoding lip movements to text, we decode text to lip movements. This similarity allows us to harness existing data sets, such as Lip Reading Sentences 3 \cite{lrs3}, used for training lip reading programs.

\section{Model Architecture}

\begin{figure}[t]
\centering
\includegraphics[width=.9\linewidth]{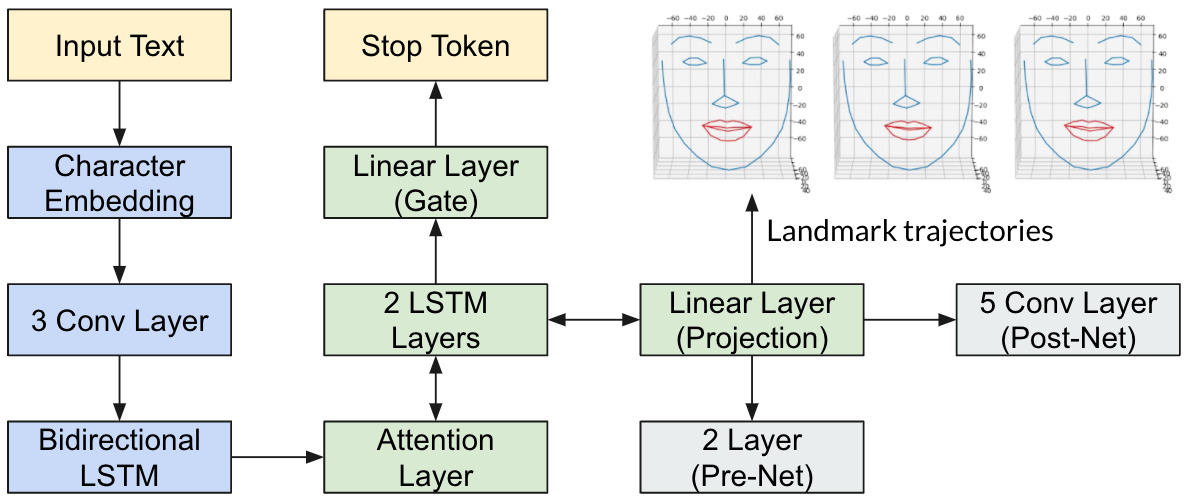}
\caption{Block diagram of Tacotron 2 architecture, with Pre-Net and Post-Net disabled for facial landmark prediction.}
\label{fig:arch}
\end{figure}

Our model architecture extends from Tacotron 2 \cite{tacotron}, a state of the art neural network for speech synthesis. Tacotron 2 uses a CNN based character level encoder, a LSTM based decoder, and an attention layer in between to help the decoder focus on specific characters during speech synthesis (see Fig. \ref{fig:arch}). The output from LSTM decoder is fed into 2 linear layers, one predicts the stop token (gate layer) while the other is used as input (Pre-Net) for the next time step. We discuss changes to the original architecture in this section.

\subsection{LSTM Decoder}

The original decoder of Tacotron 2 predicts mel spectrogram with 50 ms frame size, 12.5 ms frame hop, and 80 filterbank channels spanning 125 Hz to 7.6 kHz. We conditioned the same decoder architecture to predict 60 floating point values, representing displacements in Cartesian coordinates of 20 landmark positions around the lips. For consistency with the gate layer, we used 12.5 ms as the frame interval, giving our landmark decoder a resolution of 80 frames per second (fps).

During training, we freeze the encoder and gate layer weights and only regress on the decoder using smooth L1 loss \cite{rcnn}. Our motivation here is to reuse the encoder state pre-trained on LJ Speech 1.1 \cite{ljspeech} data set since our training set has less than 5 minutes of video. Intuitively, different sounding words are produced by different lip movements while similar sounding words are produced by similar lip movements. We examine the validity of this correlation in Section \ref{results} by comparing the best validation loss with and without freezing the encoder weights.

Unlike the original LJ Speech 1.1 data set, our training samples were extracted from multiple speakers. Therefore, we chose smooth L1 loss over mean squared error (L2 loss) so that our decoder is regressed towards the median of all speakers rather than the mean. In practice, either loss works as we did not observe any qualitative difference between predicted lip movements.

\subsection{Post-Net}

Post-Net is a 5-layer convolutional network used by Tacotron 2 to improve the overall reconstruction of mel spectrogram. It predicts a residual to be added to decoder output, where the sum is regressed towards the same target as the decoder using MSE loss. In our experiments, we noticed that the reconstructed lip movements were extremely chaotic. Removing Post-Net stabilised landmark predictions.

\subsection{Pre-Net}

Pre-Net is used by Tacotron 2 as an information bottleneck to learning attention. It contains 2 fully connected layers of 256 hidden ReLU units. At each time step, the previous step's output is preprocessed by the Pre-Net before being passed to LSTM decoder. In our experiments, removing Pre-Net significantly improved validation loss and convergence speed.

\subsection{Gate Layer}

The gate layer predicts the stop token for an input sequence and is notoriously difficult to train. Peng \cite{mmi} showed that it took 25,000 training steps for the gate layer to converge when trained on 24.6 hours of English speech using batch size 64. Given our limited sample size, we had to use pre-trained gate layer weights by aligning the animated frame interval with the frame hop duration of mel spectrogram.

\section{Experiments and Results}
\subsection{Data Preparation}

We used a small subset of the Lip Reading Sentences 3 \cite{lrs3} data set for this project. LRS 3 is compiled from a collection of TED talks uploaded on YouTube. A short clip (about 3 seconds) is extracted from each video where the speaker's front profile is clearly visible. Each clip is transcribed word for word and time synchronized with the corresponding visual frame. In addition, all numbers in this data set is spelled out, i.e. ``42'' is written as ``forty two''.

\begin{figure}[t]
\centering
\includegraphics[width=.9\linewidth]{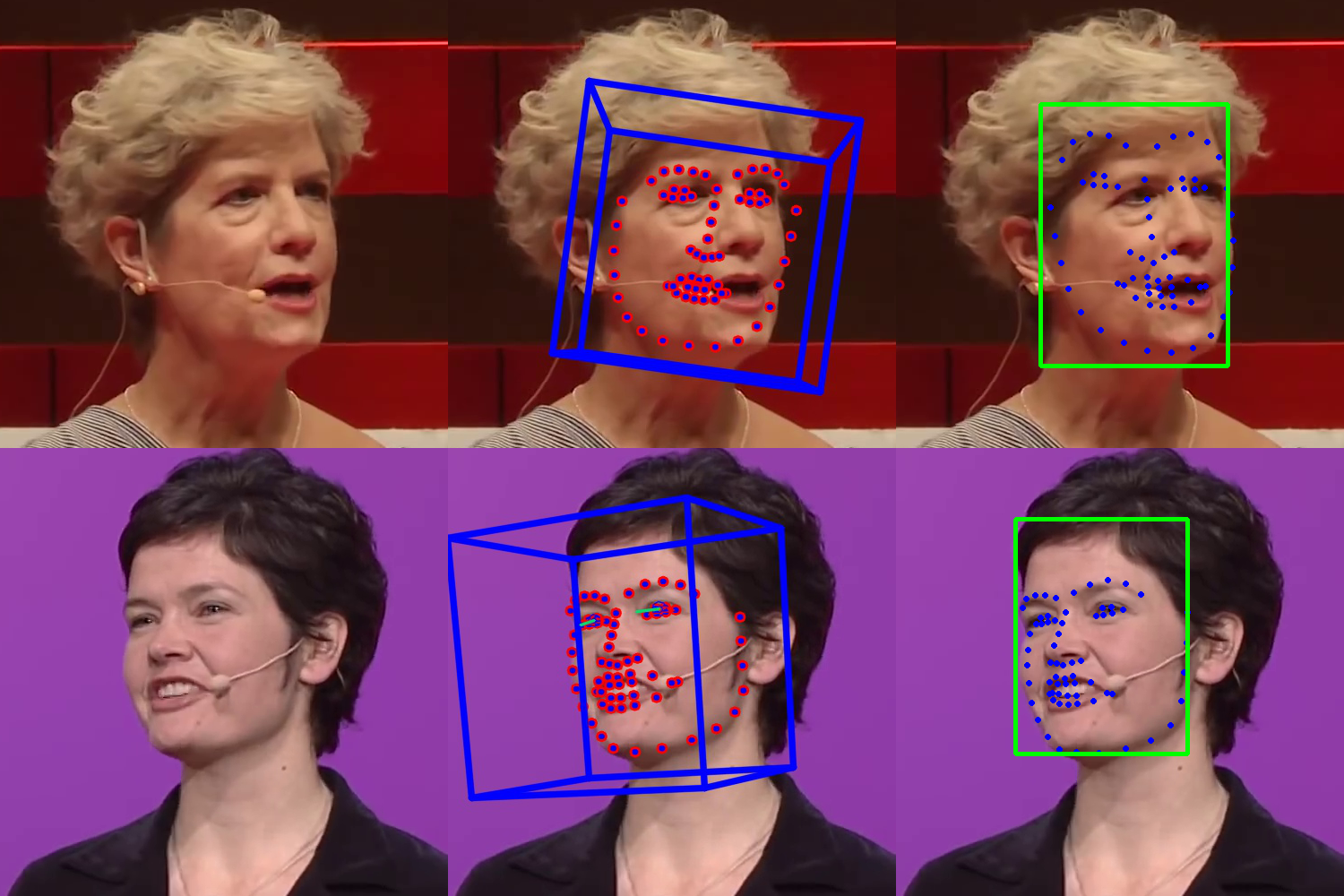}
\caption{Top row (left to right): original speaker image, misaligned facial landmarks by OpenFace, misaligned landmarks by dlib. Bottom row (left to right): original speaker image, correctly aligned facial landmarks by OpenFace, correctly aligned landmarks by dlib.}
\label{fig:misalign}
\end{figure}

Our model uses the transcript as input to predict the visual landmarks in each frame. The ground truth labels are generated by running the OpenFace 2.0 \cite{openface} algorithm on each frame of the source video. This extracts 68 facial landmark positions around the eyes, nose, mouth, and jaw line of a person. The speaker's eye gaze direction and head orientation were also computed to fit the coordinates in 3D. When training the decoder model, we used Manhattan distance between predicted and ground truth coordinates as the evaluation metric.

Most speakers wore a microphone that slightly obstructed the speaker's face. This confused OpenFace occasionally to produce misaligned facial landmarks (see Fig. \ref{fig:misalign}). Since it's costly to retrain the OpenFace model, we manually filtered out any affected videos to make a well-labelled data set. These misaligned landmarks typically have low confidence scores and we used a threshold of 0.7 to discard them.

As some videos were filmed in 2011, they suffer from degraded video quality as the resolution was limited to 480p. We used only recent videos with either 720p or 1080p resolution for this work. To account for different source video frame rates, all generated landmarks are interpolated to 80 fps.

\subsection{Label Generation}

\begin{figure}[t]
\centering
\includegraphics[width=.8\linewidth]{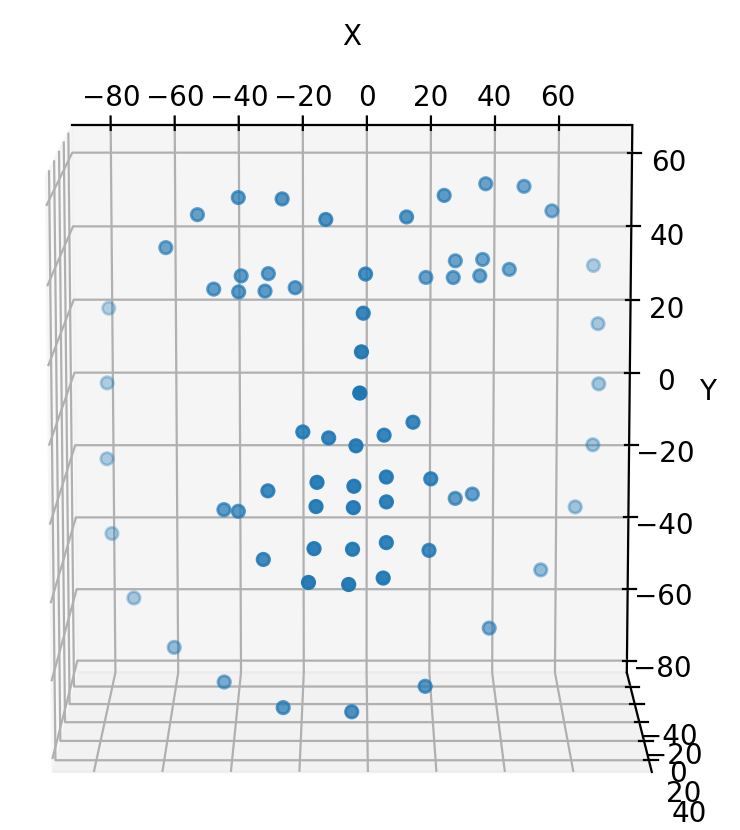}
\caption{Reprojected facial landmarks in world space.}
\label{fig:face_3d}
\end{figure}

We extract the lip positions in world coordinate space (X, Y, Z) from the output of OpenFace 2.0. The origin of the world coordinate system is the camera center, with positive axis defined by the left hand rule. Each landmark position is reprojected to the origin by a translation and a rotation using 6 estimated pose parameters (see Fig. \ref{fig:face_3d}), namely $T_x$, $T_y$, $T_z$, $R_x$, $R_y$, $R_z$. We use these reprojected coordinates as training labels to guide our model predictions to be invariant to speaker's head movement.

Since the head pose in each frame is estimated independently, there exist slight inconsistencies in the estimated camera parameters across consecutive frames. We reduce these temporal effects by selecting video clips with minimal camera movement. In addition, other effects like camera focal length, lens distortion, variances in head sizes of speakers are assumed to be negligible in this study.

To account for variations in facial features of individual speakers, we computed reference facial landmark positions for each speaker by averaging the detected coordinates of the initial frame from all clips that the speaker appears in. We then subtracted the reference positions from all coordinates in subsequent frames. This improved loss significantly as the model is trained to predict displacement from the reference frame rather than the absolute positions of facial landmarks. At inference time, the model output is added to initial landmark positions to obtain the final trajectories.

\subsection{Training Setup}
Our model is trained using Adam optimiser \cite{adam} with an initial learning rate 0.002, smooth L1 loss with beta 1.0, and batch size of 8 for 500 epochs. At each time step, teacher forcing is used to feed the correct decoder output instead of the predicted one. Each training run took 7 hours on a single Nvidia A100 GPU using approximately 8GB of memory. We computed the validation loss every 5 epochs (50 iterations) and saved the best model as a checkpoint. To achieve fast convergence, we used one-cycle LR scheduler \cite{cycle} with step size of 4000. Additionally, fp16 gradient scaler was used to improve inference speed.

While searching for the optimal hyper-parameters, we temporarily capped the maximum number of epochs to 50. We also used only 20 clips from a single speaker data set initially to further reduce training time. Once the model began to overfit, we quadrupled the size of our data set to include more speakers.

\subsection{Results} \label{results}

\begin{figure}[t]
\centering
\includegraphics[width=\linewidth]{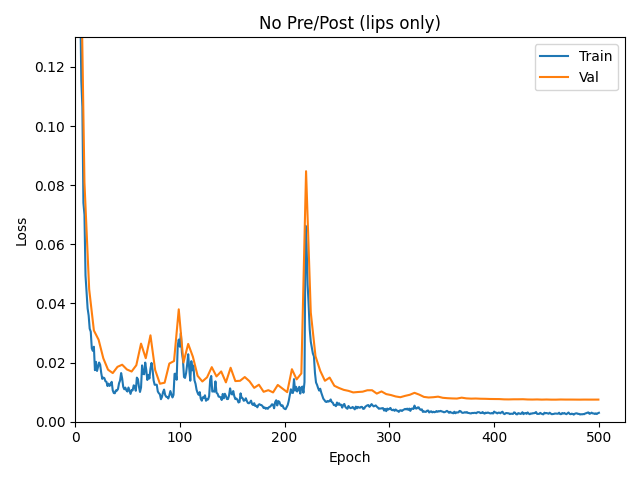}
\caption{Training and validation loss for decoder.}
\label{fig:loss}
\end{figure}

Fig. \ref{fig:loss} shows the training and validation loss for the decoder over 500 epochs. The best validation loss is achieved at the 458\textsuperscript{th} epoch with a value of $7.466\times 10^{-3}$. Compared to Nvidia's Tacotron 2 which took 50,000 iterations to train on English text, our landmark extension module converged in less than 5,000 iterations.

When trained on absolute positions instead of displacements from a reference frame, the initial loss was orders of magnitude higher. After convergence, the validation loss was also an order of magnitude higher. Our interpretation of this difference is that the model learnt more about the shape of a face rather than the movement trajectory of facial landmarks. Hence, normalisation is an important preprocessing step.

\begin{figure}[b]
\centering
\includegraphics[width=\linewidth]{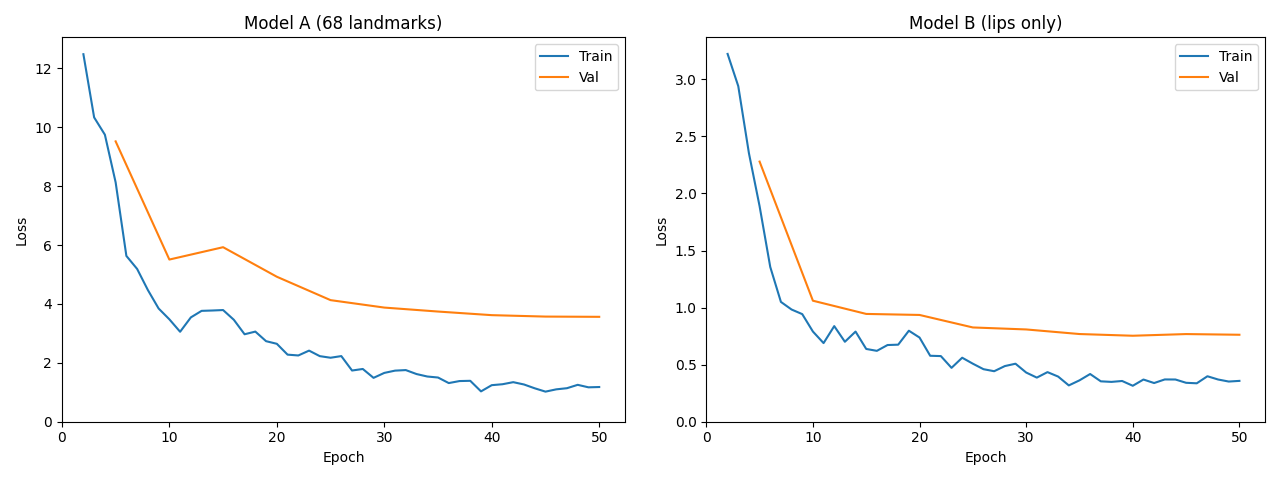}
\caption{Training and validation loss for model A and model B.}
\label{fig:comp}
\end{figure}

In addition to comparing different decoder architectures, we also experimented with predicting all 68 facial landmark positions (model A), and only predicting lip positions (model B). Fig. \ref{fig:comp} shows the training and validation loss for both models over 50 epochs. The constraint on lip positions in model B helped reduce the complexity of the task and hence is able to achieve a lower loss. An alternative is to include an additional regularization term that penalizes lip positions more. Both approaches help the model to focus on learning lip positions.

For ablation study (Table~\ref{tab:ablation}), we ran all experiments for a maximum of 50 epochs to limit resource usage. The highest validation loss occurred with adding Post-Net. This is expected because the loss function sums up the decoder loss with an additional loss term between the Post-Net output and the ground truth target. Looking at the predictions qualitatively, the output from Post-Net changes abruptly between time steps which do not represent natural speech.

\begin{table}[t]
\caption{Ablation study of Pre/Post-Net and pre-trained Encoder}
\begin{center}
\begin{tabular}{|c|c|c|c|c|}
    \hline
    \textbf{Val Loss} & \textbf{Epoch} & \textbf{Post-Net} & \textbf{Pre-Net} & \textbf{pre-trained} \\
    \hline
    $1.7127\times 10^{-1}$         & 50 & \checkmark & \checkmark & \checkmark \\
    \hline
    $7.1896\times 10^{-2}$         & 50 &            & \checkmark & \checkmark \\
    \hline
    $1.5066\times 10^{-2}$         & 50 &            &            & \checkmark \\
    \hline
    $8.9430\times 10^{-3}$         & 50 &            &            &            \\
    \hline
\end{tabular}
\label{tab:ablation}
\end{center}
\end{table}

Comparing training with and without Pre-Net, we observe that the validation loss is higher with Pre-Net, meaning it takes longer to converge and is potentially less accurate. Qualitative assessment of the inference output suggests that adding Pre-Net does not make lip movements appear more natural. Therefore, we removed it for simplicity of architecture.

The lowest validation loss results from not using pre-trained encoder weights. However, when we looked at the inference output, the generated lip landmarks did not move at all. We believe this is due to the absence of teacher forcing at inference time. Since there is nothing to correct decoder output in subsequent time steps, the whole prediction from our autoregressive model becomes misaligned if the input produced by the encoder is wrong. Given these observations, we believe transfer learning between audio and visual speech data is effective in aiding convergence.

\section{Discussion}

One limitation of our evaluation is the lack of a common quantitative benchmark for model performance. This is generally a difficult problem because there are many different, but equally valid, ways of speaking the same sentence. For example, when making a public speech, we project our voices differently than whispering into someone's ears. How to quantitatively normalise these variations in tone, context, and speaker emotion is left as an area for future work.

\subsection{Animation Quality}

\begin{figure}[b]
\centering
\includegraphics[width=\linewidth]{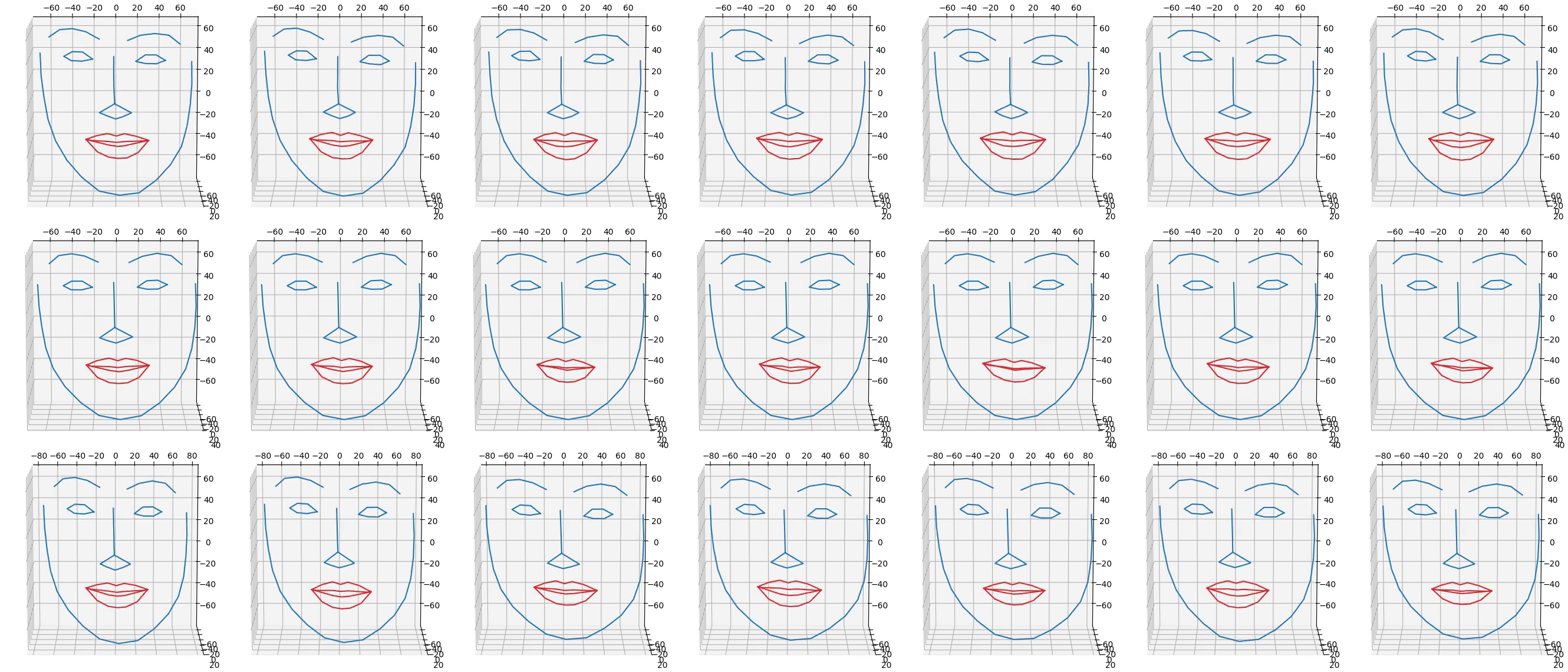}
\caption{Top row: model A regresses on all 68 landmark positions, lips don't close. Middle row: model B regresses on just lip positions, closes properly. Bottom row: ground truth positions from validation set, generated by OpenFace 2.0.}
\label{fig:quality}
\end{figure}

To study our model predictions qualitatively, we used the sentence "IT'S NOT JUST MY WORK" as input to both model A and model B for inference. We generated the frame-by-frame results using \textit{}{matplotlib} for display in Fig. \ref{fig:quality}. The lips are highlighted in red for better contrast.

Looking at the predicted trajectory of model B and ground truth positions, both lip movements look natural to the human eye. This suggests that the labels generated by OpenFace 2.0 are suitable for training landmark prediction models. Model A, on the other hand, predicted movements in other facial features like eye brows and jaw lines more than lip movements. Future work could look into refining these predictions for all 68 facial landmarks.

\begin{figure}[t]
\centering
\includegraphics[width=\linewidth]{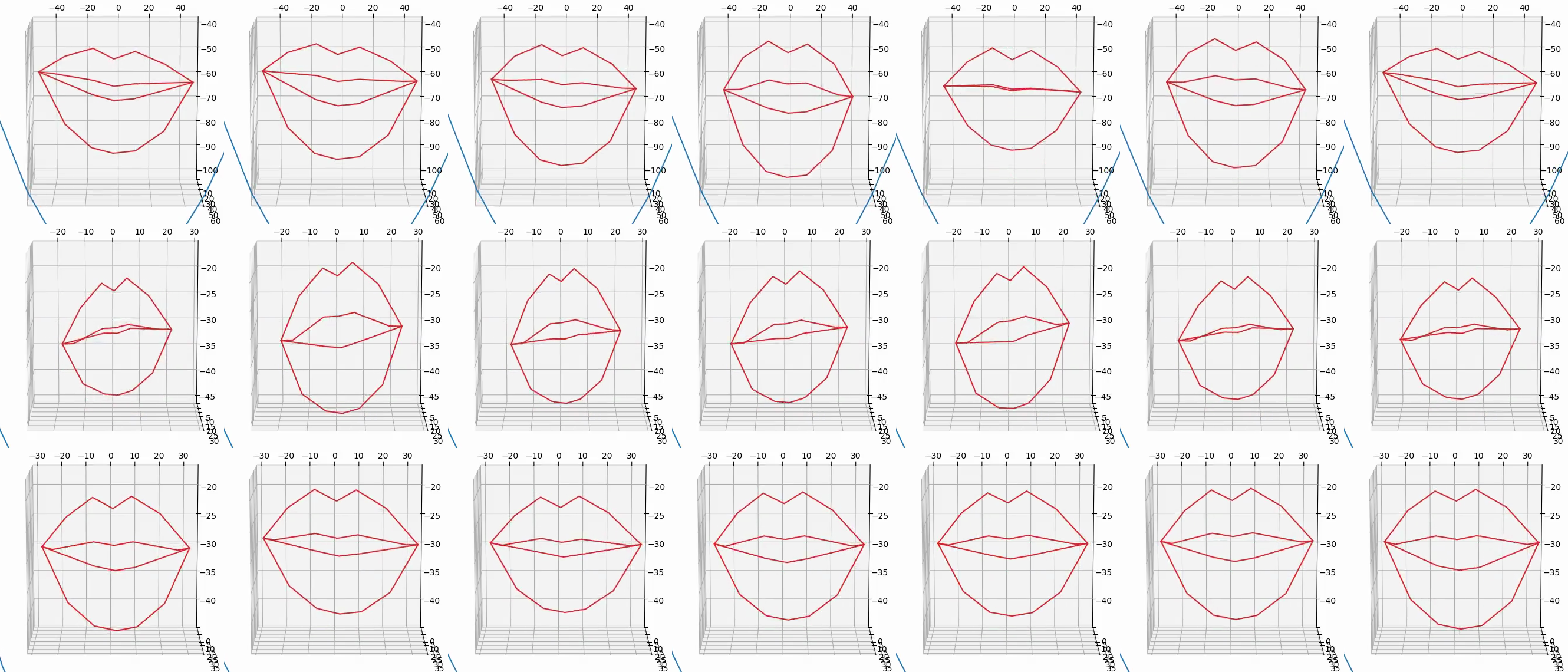}
\caption{Top row: lip movements predicted by NR 1D CNN audio-driven approach. Middle row: lip movements predicted by our text-driven approach. Bottom row: ground truth positions from validation set, generated by OpenFace 2.0.}
\label{fig:audio}
\end{figure}

Using the same sentence, we extracted WAV files from original videos to compare with existing audio-driven approach. Each audio file is converted to a 64-channel mel spectrogram with 40ms non-overlapping Hanning window before being consumed by a noise resilient 1D convolutional neural network \cite{gen3d} to output 68 facial landmarks. Fig. \ref{fig:audio} shows the differences in lip movements between audio and text-driven approaches. Due to differences in training data sets, the audio approach tends to produce rounder mouth shapes than our ground truth labels. Note that the example sentence is taken from our validation set which is not exposed to the NR 1D CNN model.

\begin{figure}[t]
\centering
\includegraphics[width=\linewidth]{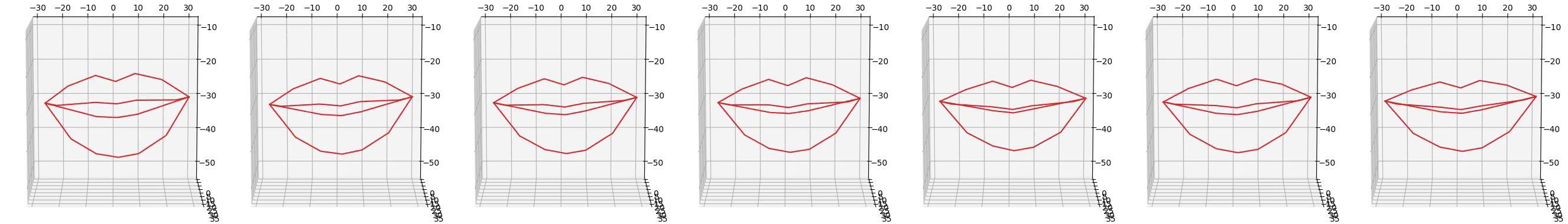}
\caption{Lip movements for pronouncing unseen text: the first 4 frames correspond to "HELLO" while the last 3 frames correspond to "WORLD".}
\label{fig:lips}
\end{figure}

To investigate how well our model generalizes to unseen input text, we feed simple tokens like "HELLO WORLD" to model B and observed its lip movements. Fig. \ref{fig:lips} shows the frame by frame visualisation of the predicted pronunciation. More work needs to be done on long and complex sentences.

As the frame-by-frame differences are sometimes too subtle to be noticed, we have uploaded video renderings of both text and audio-driven approaches together with our paper.

\subsection{Applications to Metaverse}

To help animators create more realistic avatars for metaverse, the industry has released many SDKs such as Nvidia's Omniverse Audio2Face\footnote{https://www.nvidia.com/en-us/omniverse/apps/audio2face/}, Meta's Oculus Lipsync\footnote{https://developer.oculus.com/documentation/unity/audio-ovrlipsync-unity/}, and the open-source tool uLipSync\footnote{https://github.com/hecomi/uLipSync}. These tools generate animations based on hand crafted visemes, which are expensive to create and limited in simulating the full range of emotions and facial expressions occurring in natural speech.

Using a deep learning approach, we can harness the vast number of videos uploaded to YouTube and other platforms in the wild to train more realistic facial landmark prediction models. These models capture subtle visual cues that might escape even the most experienced human animator.

While the research community has pushed the envelope in model development, the industry also needs easier ways for animators to adopt such models in their daily work. Deploying these advanced models on edge hardware, like VR headsets and other IoT devices, also bring additional concerns on power consumption and real time inference speed.

While these concerns are non-trivial to solve, we remain excited about the potential of text-to-speech/animation synthesis and its applications to metaverse in the future.

\section{Conclusion}

In conclusion, we proposed a weakly supervised approach to generating facial landmark positions directly from text input. We automated image labelling tasks by applying OpenFace 2.0 to YouTube videos. To reduce training time, we successfully reused encoder embedding trained on 24.6 hours of LJ Speech 1.1 data set. The resulting model converged quickly using just 5 minutes worth of transcribed video as training data. With improved facial landmark detectors, we can scale up to more unseen videos, allowing individuals to quickly customise the model to their own speaker profile.

For future work, we would like to explore ways to incorporate tone, context, and speaker emotion into the model \cite{disentangled}. This could allow customisation of the generated avatar based on user's profile. To further increase the training corpus and consequently the learning capacity of the model, we might want to experiment with self-supervised methods like masked autoencoders. With sufficient training data, it should be possible to expand the output keypoints to animate the entire avatar.

\section*{Acknowledgment}

We thank Oscar Chang and Vince Tan for providing feedback on this work. This research is supported in part by Nanyang Technological University Startup Grant; in part by the Singapore Ministry of Education Academic Research Fund under Grant Tier 1 RG97/20, Grant Tier 1 RG24/20 and Grant Tier 2 MOE2019-T2-1-176; in part by the NTU-Wallenberg AI, Autonomous Systems and Software Program (WASP) Joint Project; and in part by the Singapore NRF National Satellite of Excellence, Design Science and Technology for Secure Critical Infrastructure under Grant NSoE DeST-SCI2019-0012.

\bibliographystyle{IEEEtran}
\balance
\bibliography{references}
\end{document}